\documentclass{article}
\usepackage{spconf,amsmath,graphicx}
\usepackage{url}
\usepackage{booktabs}
\usepackage{amssymb}
\usepackage[font=small,labelfont=bf]{caption}

\title{DYNAMIC WEIGHT ALIGNMENT FOR TEMPORAL CONVOLUTIONAL NEURAL NETWORKS}
\name{Brian Kenji Iwana, Seiichi Uchida\thanks{This research was partially supported by MEXT-Japan (Grant No. J17H06100) and NTT Communication Science Laboratories.}}
\address{Department of Advanced Information Technology, Kyushu University, Fukuoka, Japan}
\begin{document}
%
\small
\maketitle
\begin{abstract}
In this paper, we propose a method of improving temporal Convolutional Neural Networks (CNN) by determining the optimal alignment of weights and inputs using dynamic programming. Conventional CNN convolutions linearly match the shared weights to a window of the input. However, it is possible that there exists a more optimal alignment of weights. Thus, we propose the use of Dynamic Time Warping (DTW) to dynamically align the weights to the input of the convolutional layer. Specifically, the dynamic alignment overcomes issues such as temporal distortion by finding the minimal distance matching of the weights and the inputs under constraints. We demonstrate the effectiveness of the proposed architecture on the Unipen online handwritten digit and character datasets, the UCI Spoken Arabic Digit dataset, and the UCI Activities of Daily Life dataset.
\end{abstract}
\section{Introduction}
\label{sec:intro}

Neural networks and perceptron learning models have become a powerful tool in machine learning and pattern recognition. 
Early models were introduced in the 1970s, but  recently have achieved state-of-the-art results due to improvements in data availability and computational power~\cite{schmidhuber2015deep}.
Convolutional Neural Networks~(CNN)~\cite{lecun1998gradient} in particular have achieved the state-of-the-art results in many areas of image recognition, such as offline handwritten digit recognition~\cite{wan2013regularization}, text digit recognition~\cite{lee2016generalizing,uchida2016further}, and object recognition~\cite{clevert2015fast,he2016deep}. 

Most recent successes in time series recognition have been through the use of Recurrent Neural Networks~(RNN)~\cite{jaeger2002tutorial} and in particular, Long Short-Term Memory~(LSTM) networks~\cite{Hochreiter_1997}. 
Typically, CNN-based models have been used in the image domain, however, they have also been used for time series patterns. 
A predecessor to CNNs, Time Delay Neural Networks (TDNN)~\cite{waibel1989phoneme,lang1990time} used time-delay windows similar to the filters of CNNs. 
CNNs were also used to classify time series by embedding the sequences into vectors~\cite{Zheng_2014} and matrices~\cite{razavian2015temporal,yang2015deep}.

CNNs use sparsely connected shared weights that act as a feature extractor and maintain the structural aspects from the input. 
In particular, these shared weights are linearly aligned to each corresponding window value of the input. 
However, the linear alignment assumes that each element of the input window correspond directly to each weight of the filter in a one-to-one fashion. 
It is possible that there is a more optimal alignment of the shared weights and the input values.


We propose a method of finding that alignment using dynamic programming, namely Dynamic Time Warping~(DTW)~\cite{sakoe1978dynamic}. 
DTW estimates the globally minimal distance between two time series patterns by elastically matching elements using dynamic programming along a constrained path on a cost matrix. 
While DTW is traditionally used just as a distance measure, we exploit the elastic matching byproduct of DTW to align the weights of the filter to the elements of the corresponding receptive field to create more efficient feature extractors for CNNs.

The contribution of this paper is twofold. 
First, we propose a novel method of aligning weights within the convolutional filters of CNNs by dynamically matching the weights to similar input values. 
Using the discovered dynamic weight alignment, we create a nonlinear matching to create more effective convolutions.  
Second, we demonstrate the effectiveness of the proposed method on multiple time series datasets including:  Unipen online handwritten character datasets, the UCI Spoken Arabic Digit dataset, and the UCI Activities of Daily Life dataset and perform a comparative study to reveal the benefits of the proposed weight alignment. 


\begin{figure*}
\begin{center}
\setlength\fboxsep{0pt}
\setlength\fboxrule{0pt}
\framebox[.9\columnwidth]{\includegraphics[width=.82\columnwidth]{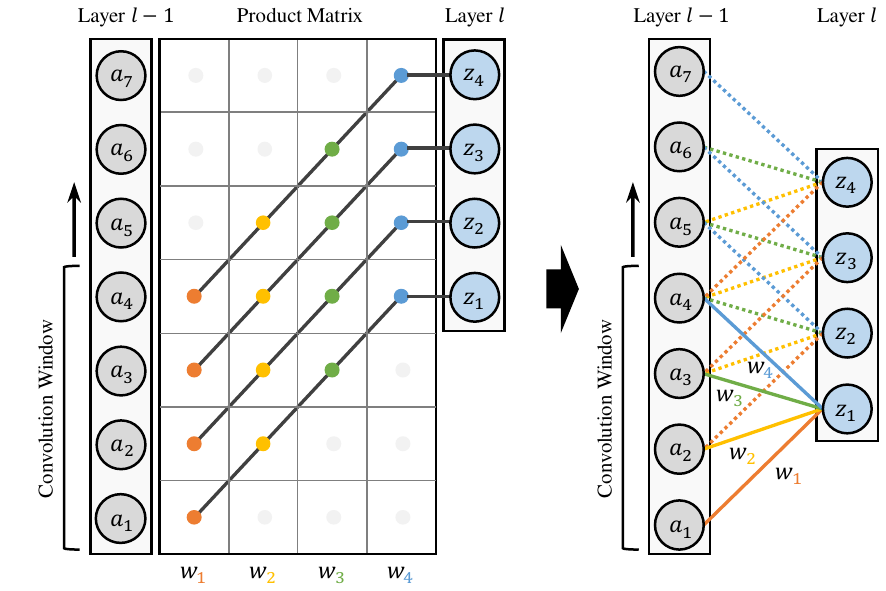}}
\framebox[.15\columnwidth]{}
\framebox[.9\columnwidth]{\includegraphics[width=.82\columnwidth]{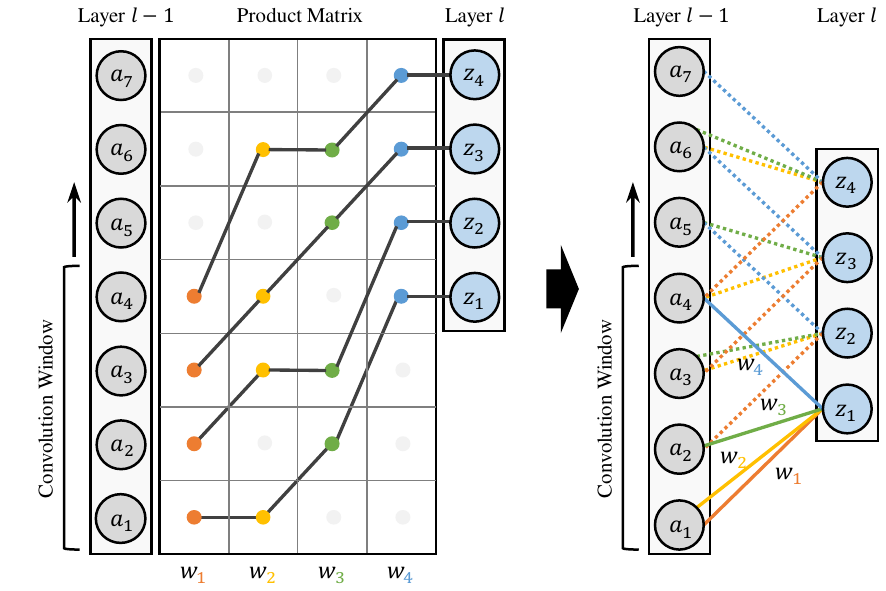}}
\framebox[.9\columnwidth]{\small (a) Convolution with linear weight alignment}
\framebox[.15\columnwidth]{}
\framebox[.9\columnwidth]{\small (b) Convolution with dynamic weight alignment}
\end{center}
\vspace{-0.4cm}
\caption{\small The comparison between a conventional linear convolution~(a) and the proposed convolution with dynamic weight alignment~(b). Both illustrate 1D convolutions with four weights $w_1,\dots,w_4$ at stride 1. The layer $l-1$ is the previous layer with elements $a_1,\dots,a_7$ and layer $l$ is the resulting feature map from the convolution with elements $z_1,\dots,z_4$. Each dot is the product of the corresponding weight and input and the blue circle is the sum of the products.}
\label{fig:cnncompare}
\end{figure*}

\section{Relation to prior work}
\label{sec:related}
Dynamic neural networks is an emerging field in neural model learning
Dynamic Filter Networks~(DFN)~\cite{de2016dynamic} use filter-generating networks to produce filters that are used depending on the input. 
Dynamic Convolutional Neural Networks~(DCNN)~\cite{kalchbrenner2014convolutional} use dynamic $k$-Max Pooling to simplify CNNs for sentence modeling. 
Deformable Convolutional Networks~\cite{dai2017deformable} use deformable convolutions to relax the constraints of a traditional convolutional window.
DTW-NNs~\cite{iwana2016robust} similarly use DTW as a nonlinear inner product for regular feed forward neural networks. 
The distinction between these models and the proposed method is that we use dynamic programming to estimate the optimal weight alignment within convolutions.

\section{Dynamic Weight Alignment for CNNs}
\label{sec:nonlinear}

The goal of the proposed method is to exploit dynamic programming to determine the optimal alignment of weights for convolutional layers in CNNs. 
In this case, we define ``optimal'' as the globally minimal warping path determined by DTW. 
In other words, instead of the conventional linear inner product of a convolution, the convolutional filter weights and the input window values are dynamically matched to minimize the difference between similar features of the weights and the input values. 
Figure~\ref{fig:cnncompare} demonstrates the difference between a conventional convolutional layer with linear weight alignment and the proposed CNN with dynamic weight alignment.

\subsection{Convolutional Neural Networks}


A CNN is an artificial neural network which contains one or more convolutional layers. 
The key features of convolutional layers is that they have sparse connectivity and use parameter sharing.
Specifically, the weights of a convolutional layer are shared for each corresponding output element's local receptive field. 
In this way, a forward calculation of a convolutional layer is identical to a convolution operation where the shared weights are the filter and the output is a feature map. 

Formally, the feature map $z^{(l)}_{j}$ of a convolutional layer is defined as:
\begin{equation}
\label{eq:conv}
z^{(l)}_{j}=\sum^{I-1}_{i=0} w^{(l)}_{i} a^{(l-1)}_{i+j}+b^{(l)}
\end{equation}
for each element $j$, where $l$ is the convolutional layer, $l-1$ is the previous layer, $i$ is the index of the filter, and $I$ is the window size. We denote $w^{(l)}_{i}$, $a^{(l-1)}_{i+j}$, and $b^{(l)}$ as the shared weights, the previous layer activations, and the bias respectively. 
In other words, $z^{(l)}_{j}$ is the inner product of the shared weights $\mathbf{w}^{l}$ and each window of the previous layer $a^{(l-1)}_{j},\dots,a^{(l-1)}_{j+(I-1)}$. 
This inner product linearly matches the weights to the inputs within the window. 
However, it is plausible that there exist instances where particular weights should be matched with more optimal inputs, for example noisy elements or feature translation and scale variance within the filter.

\subsection{Dynamic Weight Alignment}
\label{sec:weight}

The conventional inner product of a convolution acts much like a similarity function. 
Thus, the general idea is to align the weights so that there is a stronger activation to input windows that are similar but only slightly misaligned.
To optimize the alignment of weights, we adopt a dynamic programming solution, specifically DTW. 

\subsubsection{Dynamic Time Warping}
DTW is an asymmetric positive semi-definite similarity function that is traditionally used as a distance measure between sequences. 
It is calculated using dynamic programming to determine the optimal match of elements between two sequences. 
By matching elements, the sequences are \textit{warped} in the time dimension to align similar features of the time series. 

DTW finds the total cost over an optimal warping path of a local cost matrix using dynamic programming.
Given two discrete time series, sequence $\mathbf{p}=p_1,\dots,p_i,\dots,p_I$ of length $I$ and sequence $\mathbf{s}=s_1,\dots,s_j,\dots,s_J$, where $i$ and $j$ are the index of each time step and $p_i$ and $s_j$ are elements at each time step, the DTW-distance is the global summation of local distances between pairwise element matches. 
Namely, the DTW-distance is denoted as:
\begin{equation}
\label{eq:dtwdef}
\mathrm{DTW}(\mathbf{p},\mathbf{s})=\sum_{(i',j')\in\mathcal{M}}\left|\left| p_{i'} - s_{j'} \right|\right|,
\end{equation}
where $(i',j')$ is a pair of matched indices $i'$ and $j'$ corresponding to the original indices $i$ of $\mathbf{p}$ and $j$ of $\mathbf{s}$, respectively. 
The set $\mathcal{M}$ contains all matched pairs of $i'$ and $j'$.
Additionally, the set of matched pairs $\mathcal{M}$ can contain repeated and skipped indices of $i$ and $j$ from the original sequences, therefore, $\mathcal{M}$ has a nonlinear correspondence to $1,\dots,i,\dots,I$ and $1,\dots,j,\dots,J$.
$||\cdot||$ is a local distance function between elements. 


\subsubsection{Dynamic Weight Alignment with Shared Weights}

The forward pass calculation is done in two steps. 
First, DTW is calculated between the shared weights of each convolution and the receptive field window of the input. 
This is possible if we consider the weights of the convolution as the time series $\mathbf{p}$ and the window of the input as $\mathbf{s}$.  
The result is a mapping of the shared weights to the input values based on minimizing the $L^2$ distance between sequence elements.

Second, the convolution is calculated using the stored mapping. 
Namely, we propose using DTW to determine $\mathcal{M}_j$ and then calculate the result of the convolution $z^{(l)}_{j}$:
\begin{align}
z^{(l)}_{j}&=\sum_{(i', j')\in \mathcal{M}_j} w^{(l)}_{i'} a^{(l-1)}_{j'}+b^{(l)}, \label{eq:weightalignment}
\end{align}
where $\mathcal{M}_j$ is the set of matched indices $i'$ and $j'$ corresponding to the index $i$ of $\mathbf{w}^{(l)}$ and the index $j$ in $\mathbf{a}^{(l-1)}$ , respectively. 
When used in this manner, we create a nonlinear convolutional filter that acts as a feature extractor similar to using shapelets with DTW~\cite{ye2009time}.
In addition, it is important to note that unlike a conventional CNN, the set of matched indicies $\mathcal{M}_j$ allows for duplicate and skipped values of $w^{(l)}_{i'}$ and $a^{(l-1)}_{j'}$. 

The idea is that DTW will match similar features from the filter to the input and skip elements with a very high distance to the weights and perform small translations. 
Therefore, the process of aligning the weight using DTW is repeated for every stride of the convolution during all forward passes including during training and testing. 
Consequently, the alignment is only kept for the immediate forward and backward round and recalculated on the fly for subsequent iterations.

\subsection{Backpropagation of Convolutions with Dynamic Weight Alignment}

In order to train the network, Stochastic Gradient Decent~(SGD) is used to determine the gradients of the weights with respect to the error. This is done to update the weights in order to minimize the loss. 
For a CNN, the gradient of the error with respect to the shared weights is the partial derivative:
\begin{equation}
\label{eq:cnnback}
\frac{\partial C}{\partial {w^{(l)}_{i} }}=\sum_i \frac{\partial C}{\partial {z^{(l)}_{j} }} \frac{\partial {z^{(l)}_{j} }}{\partial {w^{(l)}_{i} }},
\end{equation} 
where $C$ is the loss function. 
In a conventional CNN, $w^{(l)}_{i}$ has a linear relationship to $z^{(l)}_{j}$, thus $\frac{\partial {z^{(l)}_{j} }}{\partial {w^{(l)}_{i} }}$ can be calculated simply. 
However, given the nonlinearity of the weight alignment, the calculation of the gradient is reliant on the matched elements determined by the forward pass in:
\begin{align}
\label{eq:nonlinearback}
\frac{\partial C}{\partial {w^{(l)}_{i} }}&=\sum_i \frac{\partial C}{\partial {z^{(l)}_{j} }} \frac{\partial {\left( \sum_{(i', j')\in \mathcal{M}_j} w^{(l)}_{i'} a^{(l-1)}_{j'}+b^{(l)}\right)}}{\partial {w^{(l)}_{i} }}
\\&= \delta^{(l+1)}  \sum_{(i', j')\in \mathcal{M}_j}  a^{(l-1)}_{j'}.
\end{align} 
where $\delta^{(l+1)}$ is the backpropagated error from the previous layer as determined by the chain rule.


\section{Experiments and Results}
\label{sec:results}

\subsection{Datasets and Evaluation}

We demonstrate the effectiveness of the proposed method by quantitatively evaluating the architecture and compare it to baseline methods for three diverse datasets.

The Unipen multi-writer~1a, 1b, and 1c datasets~\cite{guyon1994unipen} are constructed from pen tip trajectories of isolated numerical digits, uppercase alphabet characters, and lowercase alphabet characters respectively.  
The UCI Spoken Arabic Digit Data Set~\cite{Ganchev05comparativeevaluation} contains spoken Arabic digit patterns encoded using 13-frequency Mel-Frequency Cepstrum Coefficients~(MFCC) in 10 classes. 
The UCI Activities of Daily Life~(ADL)  Recognition with Wrist-worn Accelerometer Data Set~\cite{bruno2013analysis} is made of patterns from 7 classes of ADL actions. 
The Unipen and the UCI ADL datasets were divided into three sets for training, a test of 10\% of the data, a training set of 90\% of the data, and 50 patterns set aside from the training set for a validation set. The UCI Arabic data has a pre-defined division of the data with a speaker-independent training set and test set.

\begin{table}
\caption{\small Accuracy (\%) on the evaluated datasets. The highest accuracy for each dataset is in bold.}
\label{tab:results}
\centering
\small
\begin{tabular}{lccccc}
\toprule
&\multicolumn{3}{c}{Unipen} & \multicolumn{2}{c}{UCI}\\
\cmidrule(r){2-4} \cmidrule(r){5-6}
Method & 1a & 1b & 1c & Arabic & ADL \\
\midrule
Proposed & 98.54 & 96.08 & \textbf{95.92} & 96.95& \textbf{90.0}\\ 
\midrule
CNN & 98.08 & 94.67 & 95.33 & 95.50 & 87.1 \\
LSTM & 96.84 & 92.31 & 89.79  & 96.09& 81.4\\ 
\midrule
SVM GDTW~\cite{bahlmann2002online} & 96.2 & 92.4 & 87.9 & -- & --\\ 
HMM CSDTW~\cite{bahlmann2004writer} & 97.1 & 92.8 & 90.7 & -- & --\\
DTW-NN~\cite{iwana2016robust} & 96.8 & -- & -- & -- & --\\
Google~\cite{keysers2017multi}& \textbf{99.2} & \textbf{96.9} & 94.9 & -- & --\\
Tree Dist~\cite{hammami2010improved} & -- & -- & -- & 93.1 & -- \\
CHMM & -- & -- & -- & \textbf{98.4}   & --\\
\hfill$\Delta$($\Delta$MFCC)~\cite{hammami2012second}& &  & &  &\\
WNN~\cite{hu2011spoken} & -- & -- & -- & 96.7 & --\\
GMM + GMR~\cite{bruno2013analysis} &-- & -- & -- &--& 63.1 \\
Decision Tree~\cite{kannaactivities} &-- & -- & -- &--&80.9 \\
\bottomrule
\end{tabular}
\end{table}

\subsection{Architecture Settings}
For the experiment, we implement a five-layer CNN. 
The first two hidden layers are convolutional layers with 50 nodes of the proposed dynamically aligned filters. 
In addition, we use batch normalization~\cite{ioffe2015batch} on the results of the convolutional layers. 
The third and fourth layers are fully-connected layers with a hyperbolic tangent $\tanh$ activation and have 400 and 100 nodes respectively. 
The final output layer uses softmax with the number of outputs corresponding to the number of classes.

The learning rate $\eta_t$ at iteration $t$ is defined as $\eta_t=\frac{\eta_0}{1+\alpha t}$, where $\eta_0$ is the initial learning rate and $\alpha$ is the decay parameter. 
For all of the experiments, we use the $1/t$ progressive learning rate with a $\eta_0=0.001$ and $\alpha=0.001$ for the convolutional layers and a static learning rate of 0.0001 between the fully-connected layers.

Given that the experimental datasets are made of sequences of different dimensions, the filters should correspond accordingly. 
The convolutional filters were of size $8 \times 2$ at stride 2, $6 \times 13$ at stride 2, and $12 \times 3$ at stride 4 for the Unipen datasets, the UCI Arabic dataset, and the UCI ADL dataset, respectively. 
A stride was used to reduce redundant information and decrease computation time. 
The experiment uses batch gradient decent with a batch size of 100, 50, and 5 for the three datasets respectively and for 60,000 iterations. 
The batch sizes were selected based on the size of the training sets and were chosen to iterate through epochs at generally the same rates. 
This is the reason for the very small batch of 5 used for the ADL dataset. 

In the DTW implementation, we used the asymmetric slope constraint proposed by Itakura~\cite{itakura1975minimum} and Euclidean distance as the local distance function $||\cdot||$ of Eq.~\eqref{eq:dtwdef}. 

\subsection{Comparison Methods}

We report classification results literature as well as evaluate the datasets on established state-of-the-art neural network methods.

To evaluate the proposed method, we compare the accuracy to current methods from literature. 
For the online handwritten character evaluations, we compare results from two classical methods, SVM GDTW~\cite{bahlmann2002online} and HMM CSDTW~\cite{bahlmann2004writer}, and two state-of-the-art neural network methods, DTW-NN~\cite{iwana2016robust} and Google~\cite{keysers2017multi}. 
For the spoken Arabic digits, there is one reported neural network solution using a WNN~\cite{hu2011spoken} as well as other models using a Tree Distribution model~\cite{hammami2010improved} and a Continuous HMM of the second-order derivative MFCC~(CHMM $\Delta$($\Delta$MFCC))~\cite{hammami2012second}.
For the ADL dataset, we compare our results to the original dataset proposal~\cite{bruno2013analysis} using a Gaussian Mixture Modeling and Gaussian Mixture Regression (GMM + GMR) and the best results of Kanna et al.~\cite{kannaactivities} using a Decision Tree. 

The evaluated baselines were designed to be direct comparisons for the proposed method. 
The LSTM is used as the established state-of-the-art neural network method for sequence and time series recognition and a traditional CNN is used as a direct comparison using standard convolutional layers. 
Both comparative models are provided with the same exact training, test, and validation sets as the proposed method. 
Furthermore, the evaluated methods use the same batch size and number of iterations as the proposed method for the respective trials.
For the LSTM evaluation, an LSTM with two recursive hidden layers, two fully-connected layers, and a softmax output layer was used. 
The second comparative evaluation was using a CNN with the same exact hyperparameters as the proposed method, but with standard convolutional nodes. 


\subsection{Results and Discussion}

The results of the experiments are shown in Table~\ref{tab:results}. 
The results show that the proposed method surpassed all of the results of a conventional CNN as well as the LSTM. 
Furthermore, the results are competitive with the state-of-the-art methods despite many of them being tailored to the respective datasets and data types.






In the online handwriting and ADL experiments, the LSTM performed poorly compared to both CNNs. 
One reason for the limited performance of the LSTM is that each individual element of those datasets do not contain a significant amount of information and the model needs to know how all elements work together to form spacial structures. 
For example, for large tri-axial accelerometer data, individual long-term dependencies are not as important as the local and global structures whereas CNNs excels. 
Another reason for the poor performance of the ADL dataset could be the low amount of training data (600 training samples), high amounts of noise, and a high variation of patterns within each class.
However, the LSTM did comparatively well on the spoken Arabic digits.

\begin{figure}
\begin{center}
\setlength\fboxsep{0pt}
\setlength\fboxrule{0pt}
\framebox[.32\columnwidth]{\includegraphics[width=.33\columnwidth]{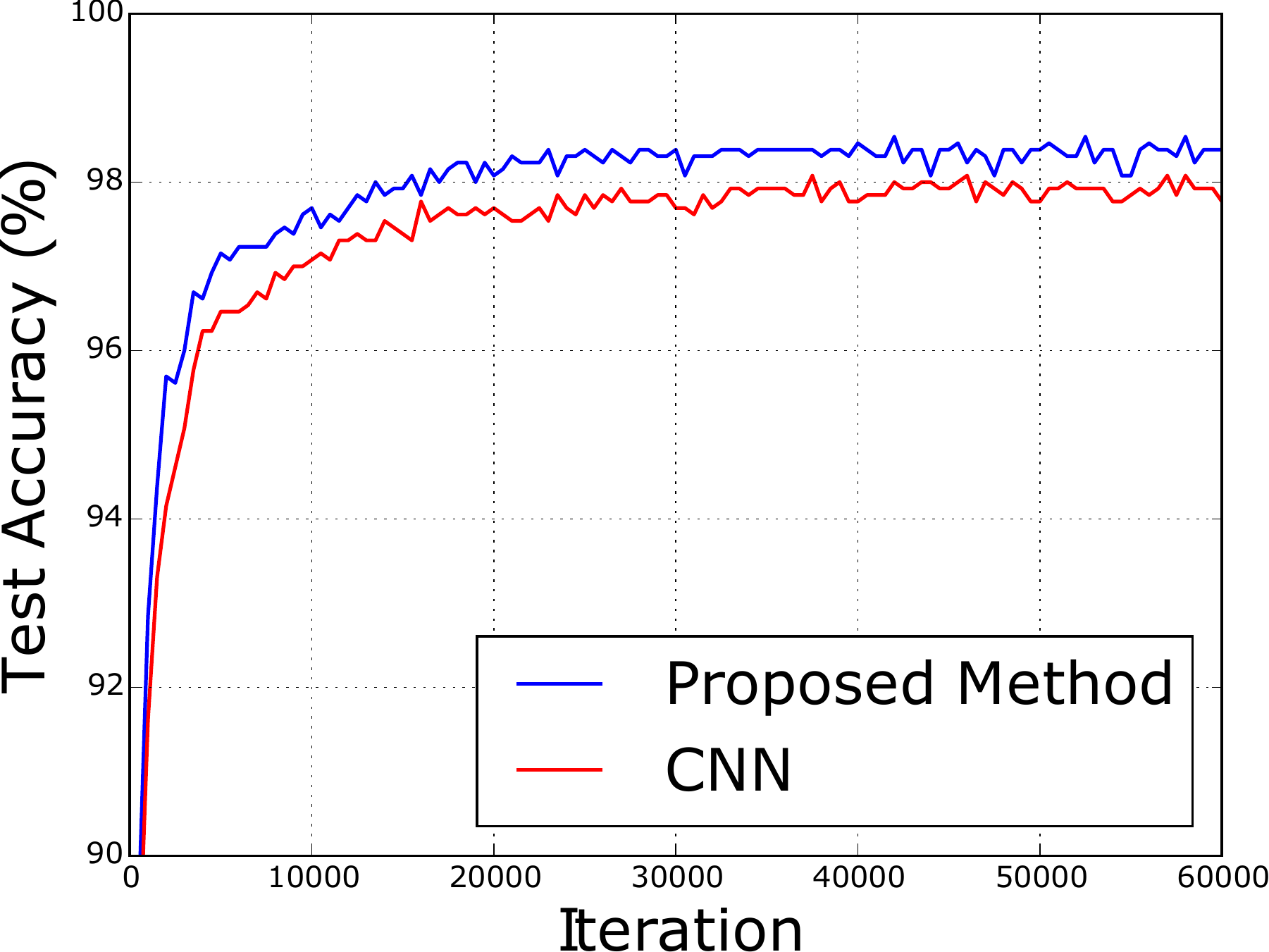}}
\framebox[.32\columnwidth]{\includegraphics[width=.33\columnwidth]{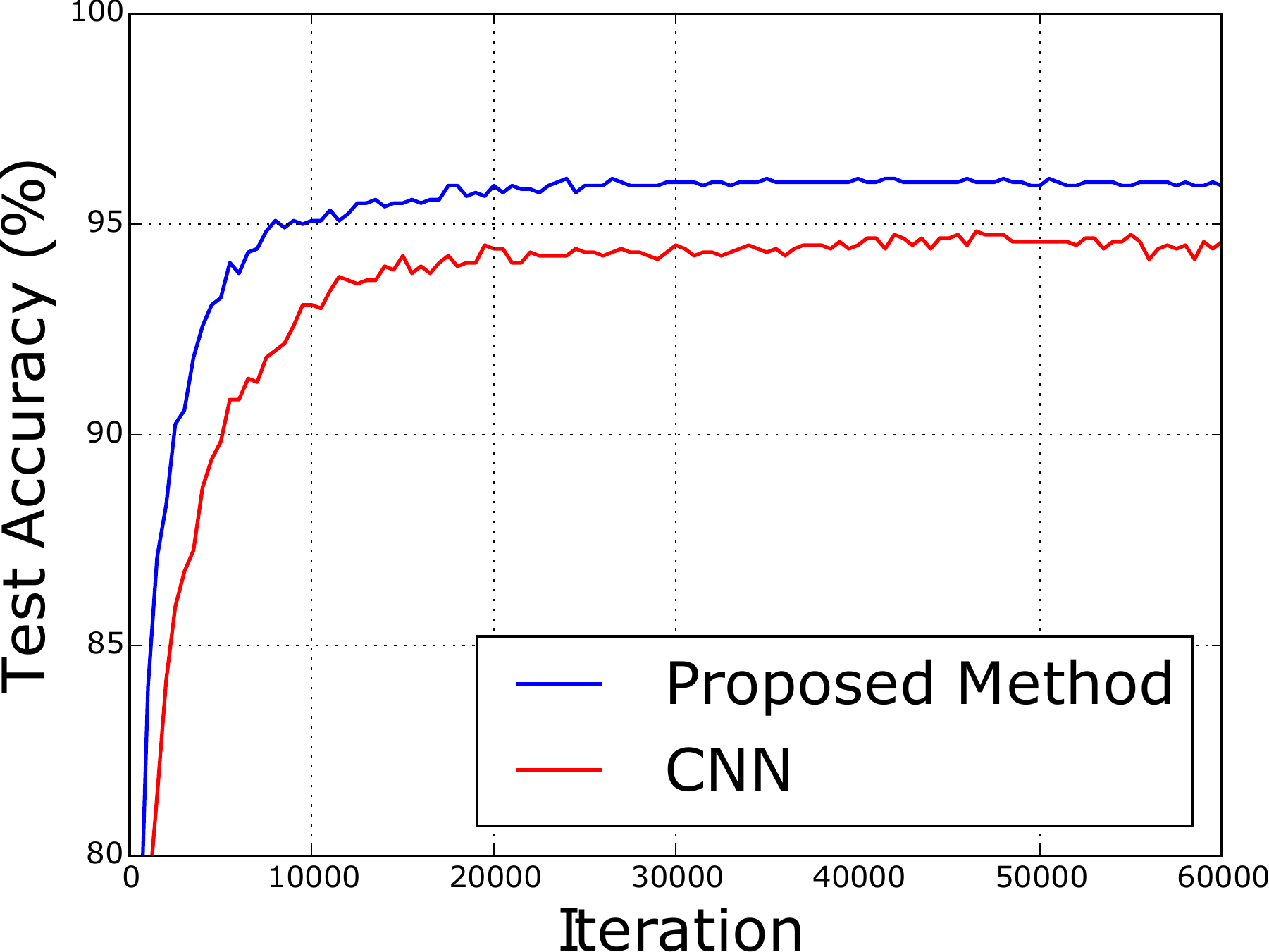}}
\framebox[.32\columnwidth]{\includegraphics[width=.33\columnwidth]{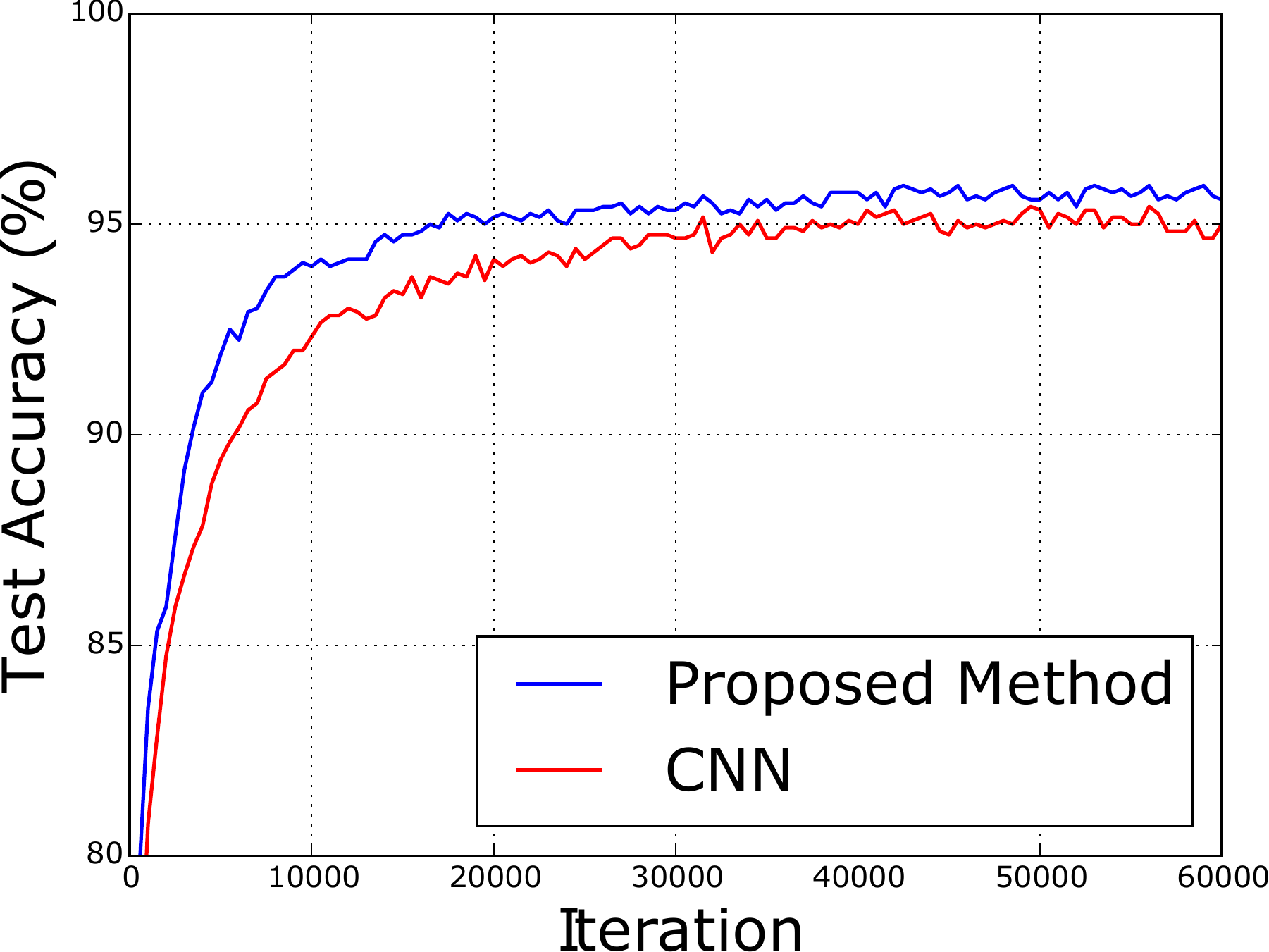}}
\framebox[.32\columnwidth]{\small Unipen 1a}
\framebox[.32\columnwidth]{\small Unipen 1b}
\framebox[.32\columnwidth]{\small Unipen 1c}
\end{center}
\vspace{-0.4cm}
\caption{\small Test accuracies of the Unipen online handwrittenof a conventional CNN and the proposed CNN with dynamic weight alignment.}
\label{fig:results}
\end{figure}

The most important comparison is the conventional CNN with linearly aligned weights against the proposed method with dynamically aligned weights. 
In addition to the increased accuracy, we observed from Fig.~\ref{fig:results} that compared to the conventional CNN, the proposed method achieves a higher accuracy during all parts of training but especially during the early stages.
This indicates that the nonlinear alignment is able to optimize the weights efficiently.

One explanation of the improved accuracy is that aligning the weights to their similar corresponding inputs is more efficient than conventional linear matching. 
The weights of a convolutional layer learned by a CNN act like filter for feature extraction~\cite{lecun1998gradient}. 
The purpose of using dynamically aligned weights is to warp the assignment of weights to their most similar corresponding inputs. 
In this way, noisy input values can be skipped and normally muted but relevant features are enhanced. 
This provides a more robust convolution. 

\subsection{Computational Complexity}

In the case of the proposed method, the number of elements in the aligned sequences is equal to $I$ and $J$, where $I$ and $J$ is the width of the filter and the input, respectively. 
Furthermore, the complexity of each DTW calculation is $O(IJ)$, which is required for every application of a convolutional filter. 
Thus, the computational complexity of the convolutional layer with dynamic weight alignment becomes $O(\frac{N  I J^2}{S})$, where $N$ is the number of convolutional nodes and $S$ is the stride. 
Compared with the standard convolution of a temporal CNN with a complexity of $O(\frac{N I J}{S})$, this a relatively small increase in complexity compared to the overall network. 

The per classification runtime for the traditional CNN was 0.036s, 0.092s, and 0.029s for the Unipen, ADL, and Spoken Arabic datasets, respectively. 
The proposed method had runtimes of 0.114s, 0.403s, and 0.078s, respectively. 
The networks were constructed in Python using Numpy with no GPU on a desktop computer with an Intel Xeon 2.6 GHz CPU. 
However, these speeds can be further optimized with the use of GPUs and deep learning libraries.

\section{Conclusion}
\label{sec:conc}

In this paper, we proposed a novel method of optimizing the weights within a convolutional filter of a CNN through the use of dynamic programming. 
We implemented DTW as a method of sequence element alignment between the weights of a filter and the inputs of the corresponding receptive field. 
In this way, the weights of the convolutional layer are aligned to maximize their relationship to the data from the previous layer. 
Furthermore, we show that the proposed model is able to tackle time series pattern recognition. 
We evaluated the proposed model on a variety of datasets to reach state-of-the-art results.
This shows that the proposed method a viable feedforward neural network model for time series recognition and an effective method of optimizing the convolutional filter in CNNs. 
There is potential for this work to be extended to any CNN-based model. 


\newcommand{\BIBdecl}{\setlength{\itemsep}{0.48mm}}
\small
\bibliographystyle{IEEETran}
\bibliography{master,deep,dtwnn}

\end{document}